\documentclass{article}

\usepackage{microtype}
\usepackage{graphicx}
\usepackage{subfigure}
\usepackage{booktabs} 

\usepackage{hyperref}

\usepackage[accepted]{icml2025}

\usepackage{amsmath}
\usepackage{amssymb}
\usepackage{mathtools}
\usepackage{amsthm}

\usepackage[capitalize,noabbrev]{cleveref}

\theoremstyle{plain}
\newtheorem{theorem}{Theorem}[section]

\theoremstyle{definition}
\newtheorem{definition}[theorem]{Definition}

\theoremstyle{remark}

\usepackage{multicol}

\newcommand{\mE}{\mathbb{E}}
\newcommand{\mU}{\mathbf{U}}
\newcommand{\mX}{\mathbf{X}}
\newcommand{\meps}{\boldsymbol{\varepsilon}}

\newcommand{\bmu}{\boldsymbol{\mu}}

\usepackage[textsize=tiny]{todonotes}

\icmltitlerunning{Estimating Causal Effects in Gaussian Linear SCMs with Finite Data}

\begin{document}

\twocolumn[
\icmltitle{Estimating Causal Effects in Gaussian Linear SCMs with Finite Data}



\icmlsetsymbol{equal}{*}

\begin{icmlauthorlist}
\icmlauthor{Aurghya Maiti}{equal,yyy}
\icmlauthor{Prateek Jain}{equal,yyy}
\end{icmlauthorlist}

\icmlaffiliation{yyy}{Department of Computer Science, Columbia University NY 10025}

\icmlcorrespondingauthor{Aurghya Maiti}{aurghya@cs.columbia.edu}
\icmlcorrespondingauthor{Prateek Jain}{prateek.jain@columbia.edu}

\icmlkeywords{Machine Learning, ICML}

\vskip 0.3in
]



\printAffiliationsAndNotice{\icmlEqualContribution} 

\begin{abstract}
Estimating causal effects from observational data remains a fundamental challenge in causal inference, especially in the presence of latent confounders. This paper focuses on estimating causal effects in Gaussian Linear Structural Causal Models (GL-SCMs), which are widely used due to their analytical tractability. However, parameter estimation in GL-SCMs is often infeasible with finite data, primarily due to overparameterization. To address this, we introduce the class of Centralized Gaussian Linear SCMs (CGL-SCMs), a simplified yet expressive subclass where exogenous variables follow standardized distributions. We show that CGL-SCMs are equally expressive in terms of causal effect identifiability from observational distributions and present a novel EM-based estimation algorithm that can learn CGL-SCM parameters and estimate identifiable causal effects from finite observational samples. Our theoretical analysis is validated through experiments on synthetic data and benchmark causal graphs, demonstrating that the learned models accurately recover causal distributions.
\end{abstract}
\section{Introduction}

Identifying causal effects from observational data is a foundational challenge in causal inference, largely due to the pervasive issue of unobserved confounding. Over the past decades, significant progress has been made in addressing this challenge in the non-parametric setting, culminating in robust theoretical frameworks such as Pearl’s do-calculus, and systematic algorithmic approaches for identification \cite{pearl1995causal, tian2002general, shpitser2012counterfactuals, huang2012pearl, bareinboim2012causal, bareinboim2016causal, correa2017causal, lee2019gid, correa2021nested, lee2020general, correa2024ctfcalc}. In particular, the problems of identifying both interventional distributions ($L_2$ queries) and counterfactual distributions ($L_3$ queries) have been extensively explored and largely resolved in the non-parametric domain.

By contrast, the fields of econometrics and statistics have traditionally approached causal inference under parametric assumptions, focusing on the identifiability of causal effects in structured models—most notably in linear systems \cite{BritoPearl2002, Tian2005, chen2017identification, KumorChenBareinboim2019, kumor2020efficient, shimizu2014lingam}. However, many of these works assume either access to infinite data, partial knowledge of distributional parameters, or the Markovian assumption, which excludes latent confounding and can be overly restrictive in practical settings. On the other hand, recent efforts have aimed to estimate causal effects from finite data, but these have primarily focused on non-parametric models \cite{jung2021estimating1, jung2021estimating2}.

This work seeks to bridge the gap between these two perspectives by addressing the problem of estimating identifiable causal effects in Gaussian Linear Structural Causal Models (GL-SCMs) using only finite observational data. GL-SCMs are widely used for their interpretability and analytic convenience, but they often suffer from high parameter complexity—particularly when modeling both observed and latent confounding—making accurate parameter estimation infeasible from limited samples. 

To address this, we introduce the class of Centralized Gaussian Linear SCMs (CGL-SCMs), a simplified yet expressive subclass of GL-SCMs in which all exogenous variables are standardized to have zero mean and unit variance. This centralization significantly reduces the number of free parameters and enables tractable learning from finite data. Our key contributions are twofold:

\begin{enumerate}
    \item We prove that CGL-SCMs retain the same expressive power as GL-SCMs with respect to identifying causal effects from observational distributions (Th.~\ref{th:id-gl-scm}).
    \item Second, we develop a scalable Expectation-Maximization (EM) algorithm to estimate the parameters of CGL-SCMs from finite data (Alg.~\ref{alg-1} and Alg.~\ref{alg-2}), even in the presence of latent variables, which can then be used to identify causal effects.
\end{enumerate}

\section{Centralized Gaussian Linear SCMs}

In this section, we first investigate the class of Gaussian Linear SCMs and then introduce a new class of Centralized Gaussian Linear SCMs which have the same expressive power for effect identification from observation. In a Gaussian Linear SCM, the exogenous variables are independently sampled from normal distributions, and endogenous variables are a linear combination of their parents variables. Next, we provide a formal definition for such a model.

\begin{definition}[Gaussian Linear SCM]
A $\texttt{GL-SCM}$ is defined by the tuple $\langle (\boldsymbol{U', \varepsilon'}), \mathbf{X}, \mathbb{P}, \mathbf{F}_\mathbf{X} \rangle$ where
(i) $\mathbf{U'} \sim \mathcal{N}(\boldsymbol{\mu_{U'}}, \boldsymbol{\Sigma^2})$ is the set of exogenous confounders ($\boldsymbol{\Sigma^2}$ is a diagonal matrix)
(ii) $\boldsymbol{\varepsilon'} \sim \mathcal{N}(\boldsymbol{\mu_{\varepsilon'}}, \boldsymbol{\Psi^2})$ is the  set of exogenous non-confounding variables
(iii) The endogenous variables $X_i$ are then given by
\[
    X_i = \sum_{X_j \in Pa^o(X_i)} \alpha_{ji} X_j + \sum_{U'_k \in Pa^u(X_i)} \alpha_{ki}' U_k' + \mu_{i}' + \varepsilon_i'
\]
where $Pa^o(X_i)$ are the observed parents of $X_i$, and $Pa^u(X_i)$ are the unobserved confounding parents of $X_i$. $\mu_{i}'$ are the bias for $X_i$.
\end{definition}

Estimation of the exact parameters in such a model can be hard, and often impossible, since the number of parameters can be too large compared to the information available from the observed data. In the following part of the section, we introduce Centralized Linear SCMs, in which the exogenous variables are only sampled from standard distribution $\mathcal{N}(0, 1)$ and the errors are sampled from normal distributions with zero-mean. Such models can have much lesser number of parameters compared to GL-SCMs and these parameters can in many cases be estimated from data alone.

\begin{definition}[Centralized Gaussian Linear SCM]
A $\texttt{CGL-SCM}$ is defined by the tuple $\langle (\boldsymbol{U, \varepsilon}), \mathbf{X}, \mathbb{P}, \mathbf{F}_\mathbf{X} \rangle$ where
(i) $\mathbf{U}$: set of exogenous confounders and $\mathbf{U} \sim \mathcal{N}(\mathbf{0}, \mathbf{I})$
(ii) $\boldsymbol{\varepsilon}$: set of exogenous non-confounding variables, and $\boldsymbol{\varepsilon} \sim \mathcal{N}(\mathbf{0}, \boldsymbol{\Psi^2})$
(iii) The endogenous variables $X_i$ are then given by
\[
    X_i = \sum_{X_j \in Pa^o(X_i)} \alpha_{ji} X_j + \sum_{U_k \in Pa^u(X_i)} \alpha_{ki} U_k + \mu_{i} + \varepsilon_i
\]
where $Pa^o(X_i)$ are the observed parents of $X_i$, and $Pa^u(X_i)$ are the unobserved confounding parents of $X_i$. $\mu_{i}$ are the bias for $X_i$.
\end{definition}

A natural question now is how much do we lose from GL-SCM to CGL-SCM is there any model which can be represented in GL-SCM, that cannot be modeled in CGL-SCM? Is there an effect that can be identified in GL-SCM with graphical assumptions and observational data, but not in CGL-SCM. The following theorem shows that as far as observational distributions are concerned, both GL-SCM and CGL-SCM has the same expressive power. 

\begin{theorem}[Expressivity of CGL-SCM]
For any model $M'$ in \texttt{GL-SCM}, there exists a model $M$ in \texttt{CGL-SCM} with same causal graph such that $P^{M'}(\mathbf{X}) = P^{M}(\mathbf{X})$.
\end{theorem}

An advantage of CGL-SCM is that it can be estimated from data when GL-SCM cannot be estimated because of the higher number of parameters. Next, we state the most important observation that follows from the above discussion. Suppose a query $Q$ is identifiable in the causal graph from observational data in the GL-SCM $M'$. Now, by the definition of identifiabiity, it implies that any SCM with the same causal graph and $P(\mathbf{X})$ will result in the same result for $Q$. Hence, estimating $Q$ in the CGL-SCM with the same $P(\mathbf{X})$ gives us an estimate of $Q$ in the GL-SCM, even though the exact parameters of the later are impossible to know from the available data. We formally state the theorem as follows:

\begin{theorem}[Identification in GL-SCM] \label{th:id-gl-scm}
Let $M'$ and $M$ be a GL-SCM and CGL-SCM respectively, with the same causal graph $G$ and $P^{M'}(\mathbf{X}) = P^{M}(\mathbf{X})$. If a query $Q$ is identifiable in causal graph $G$, then
\begin{equation}
    P^{M'}(Q) = P^{M}(Q)
\end{equation}
\end{theorem}

All proofs and derivations are in the Appendix.
\section{Estimation of \texttt{CGL-SCM} using Finite Data}

For our problem, we assume that the causal graph is known and that we are given finite samples from an observational ($L_1$) distribution. In this section, we present a procedure to estimate the parameters of a CGL-SCM using only finite observational data. It is important to note that the parameter estimation problem in this setting may not have a unique solution. However, our goal is not necessarily to recover the true data-generating model, but rather to find a CGL-SCM that induces the same $L_1$ distribution. Once such a model is estimated, any identifiable counterfactual or interventional query can be computed using this estimated SCM. By Theorem~\ref{th:id-gl-scm}, the results of these queries will match those from the original model that generated the data.

To estimate the parameters of a CGL-SCM, we adopt the Expectation-Maximization (EM) algorithm, a well-established method for maximum likelihood estimation in the presence of latent variables. 
However, to apply EM in our setting, we must first reformulate the CGL-SCM in a vectorized form that aligns with the classical linear-Gaussian framework.

\subsection{CGL-SCM in Vectorized Form}
Given a CGL-SCM $M$, we first define a matrix representation of its causal structure. Let $\mathbf{X}$ denote the vector of endogenous variables, and let $G$ be the known causal graph over $\mathbf{X}$. Construct a $|\mathbf{X}| \times |\mathbf{X}|$ matrix $T$ where each entry $(i, j)$ is defined as:
\begin{equation}
    t_{ij} = 
    \begin{cases}
        \alpha_{ij} \quad \text{if causal edge $X_i \rightarrow X_j$ exists in $G$} \\
        0 \qquad \text{otherwise }
    \end{cases}
\end{equation}

Next, define the matrix $B$ to capture the total influence of one variable on another through all directed paths in the graph. Let $d$ be the length of the longest path in $G$. The matrix $B$ is computed as:

\begin{equation}
B = I + \sum_{i=1}^d T^i    
\end{equation}

This sum includes the identity matrix to account for self-influence, which we define to have unit weight. The $(i, j)$-th entry of $B$ thus represents the total (additive) influence of $X_i$ on $X_j$ across all directed paths in the graph.

In addition to $B$, define a $|\mathbf{U}| \times |\mathbf{X}|$ matrix $C$, where each entry $(i, j)$ gives the direct effect of exogenous variable $U_i$ on endogenous variable $X_j$.

Following this, the endogenous variables $\mathbf{X}$ in a $\texttt{CGL-SCM}$ can be written in terms of only $\mathbf{U}$ and $\boldsymbol{\varepsilon}$ in the vectorized form as follows:

\begin{equation}
    \mathbf{X} = B^T \boldsymbol{\mu} + B^TC^T \boldsymbol{U} + B^T \boldsymbol{\varepsilon}
\end{equation}

\begin{algorithm}[t!]
\caption{\texttt{CGL-Go}} \label{alg-1}
\begin{algorithmic}[1]
  \STATE \textbf{Initialization:} Initialize $B, C, \boldsymbol{\mu}$ and $, B_m, C_m$ as defined in \ref{subsection:algorithm}
  \STATE $B = I + B \cdot B_m$, $\quad C = C \cdot C_m$
  \REPEAT
    \STATE \textbf{E-step (Expectation):}
    \[
    \mu_{U^i \mid X^i} = CB((CB)^TCB + B^TB)^{-1}(\mathbf{X}^i - \mathbf{B}^T\boldsymbol{\mu})
    \]
    \[
    \Sigma_{U^i \mid X^i} = I - CB((CB)^TCB + B^TB)^{-1}(CB)^T
    \]
    \STATE \textbf{M-step (Maximization):}
    \STATE Maximize $l(\mathbf{B}, \mathbf{C}, \boldsymbol{\mu})$ given by
    \begin{align*}
    & - n \log|\mathbf{B}^T \mathbf{B}| \\
    & - \sum_{i=1}^N \mathbb{E}_{U^i|x^i}[(\mathbf{X}^i - (B^T\boldsymbol{\mu} + (CB)^T\mathbf{U}^i))^T \\
    & \qquad (B^TB)^{-1}(\mathbf{X}^i - (B^T\boldsymbol{\mu} + (CB)^T\mathbf{U}^i))]
    \end{align*}
    \STATE Update $\mathbf{B}$:
    \REPEAT
    \FOR{each training example $\mathbf{X}^i$}
        \STATE Compute the gradient: $\nabla_B l(B, C, \boldsymbol{\mu})$
        \STATE Update parameters: $B \gets B + \eta \nabla_B l(B, C, \boldsymbol{\mu}) \cdot B_m$
    \ENDFOR
    \UNTIL{Convergence}
    \STATE Update $\mathbf{C}$:
    \REPEAT
    \FOR{each training example $\mathbf{X}^i$}
        \STATE Compute the gradient: $\nabla_C l(B, C, \boldsymbol{\mu})$
        \STATE Update parameters: $C \gets C + \eta \nabla_C l(B, C, \boldsymbol{\mu}) \cdot C_m$
    \ENDFOR
    \UNTIL{Convergence}
    \STATE Update $\boldsymbol{\mu}$:
    \[
        \boldsymbol{\mu} = \frac{1}{N} \sum_{i=1}^N\Big((B^T)^{-1}\mathbf{X}^i - C^T \boldsymbol{\mu}_{U^i \mid X^i}\Big)
    \]
  \UNTIL{Convergence}
\end{algorithmic}
\end{algorithm}

\subsection{Algorithm}\label{subsection:algorithm}

In the E-step, we compute the distribution of $\mU^i \mid \mathbf{x}^i; \bmu, B, C \sim \mathcal{N}\big(\bmu_{\mU^i \mid \mathbf{x}^i}, \Sigma_{\mU^i \mid \mathbf{x}^i}\big)$ where,
\[
    \bmu_{\mU^i \mid \mathbf{x}^i} = CB((CB)^TCB + B^TB)^{-1}(\mathbf{x}^i - B^T\bmu)  
\]
\[
    \Sigma_{\mU^i \mid \mathbf{x}^i} = \mathbf{I} - CB((CB)^TCB + B^TB)^{-1}B^TC^T
\]

In the M-step, we want to maximize
\begin{align*}
max_{B,C, \mu} & - n \log|B^TB| \\
& - \sum_{i=1}^N\mE_{\mU^i|\mathbf{x}^i}\Big[\big(\mathbf{x}^i - B^T\bmu - B^TC^T\mU^i\big)^T \\
& \qquad \big(B^TB\big)^{-1} \big(\mathbf{x}^i - B^T\bmu - B^TC^T\mU^i\big)\Big]
\end{align*}

\begin{algorithm}[t!]
\caption{\texttt{CGL-Edge}} \label{alg-2}
\begin{algorithmic}[1]
  \FOR{each node $X_{i}$ in $\mathbf{X}$}
        \STATE Identify a topological order $\tau_i$ of endogenous variables $\Bar{X_{i}}$ connected by direct outward edge from $X_{i}$ using causal graph $G$
        \FOR{each $X_{k}$ in $\Bar{X_{i}}$ in topological order $\tau_i$}
        \STATE Calculate edge weight $t_{ik}$
            \begin{align*}
                t_{ik} = b_{ik} - \sum_{\tau_i(X_{j}) < \tau_i(X_{k})}t_{ij}b_{jk}
            \end{align*}
        \ENDFOR
    \ENDFOR
\end{algorithmic}
\end{algorithm}

We apply EM Algorithm for estimating $B, C$ and $\boldsymbol{\mu}$. Note that naively applying EM Algorithm may result in a $B, C$ which does not satisfy the graphical constraints. In order to do that we define masks $B_m$ and $C_m$. Let $T_m$ be a matrix where the element $(i, j)$ is $1$ is there is an edge $X_i \rightarrow X_j$. Then $B_m$ is defined as 
\[
A = \sum_{i=1}^d T^i_m, \qquad b^m_{ij} =
\begin{cases}
    1 \quad \text{ if } a_{ij} > 0 \\
    0 \quad \text{ otherwise }
\end{cases}
\]
where $a_{ij}$ and $b^m_{ij}$ are the element $(i,j)$ of $A$ and $B_m$ respectively and $d$ is the length of the longest path in the graph. In $C_m$, the element $(i, j)$ is $1$ if there is an edge $U_i \rightarrow U_j$ otherwise $0$.  We did not find any closed form solution for $B, C$ during maximization so we used gradient ascent to find the matrices during this step. While optimizing using gradient ascent we mask the gradients so that graphical constraints are maintained. The final algorithm is presented in Algorithm \ref{alg-1}. The derivation is shown in Appendix.

Matrix $C$ directly provides edge weight from each exogenous confounder in $\mathbf{U}$ to endogenous variables $\mathbf{X}$. But matrix $B$ only provides aggregated weight between each pair of endogenous variables. In order to identify matrix $T$ we use Algorithm \ref{alg-2}.

\section{Experiments}
\begin{figure}[t!]
    \centering
    \includegraphics[scale=0.8]{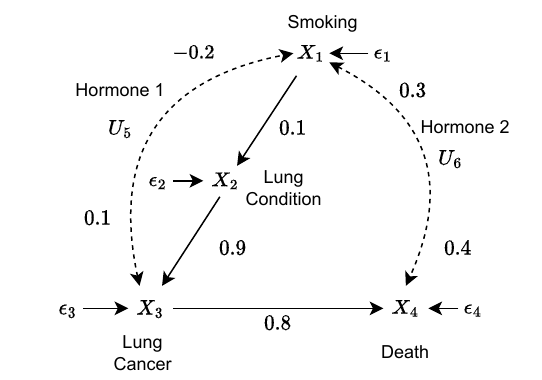}
    \caption{Napkin Causal Graph with edge weights}
    \label{fig:napkin-graph}
\end{figure}

We evaluate our algorithm on synthetic datasets generated from two synthetic Gaussian Linear SCMs. The causal graphs, known as Frontdoor Graph (Fig.~\ref{fig:frontdoor-graph}) and Napkin Graph (Fig.~\ref{fig:napkin-graph}) are widely studied in the causal inference literature. For each graph, we generated 10,000 samples of endogenous variables using its associated $\texttt{CGL-SCM}$ and applied Alg.~\ref{alg-1} and~\ref{alg-2} to estimate the SCM parameters from observational data. We then compared the resulting causal distributions with those of the original models. Full parameter matrices are provided in the Appendix; representative comparisons are shown below in Tab.~\ref{tab:frontdoor-table} and ~\ref{tab:napkin-table}.

\subsection{Frontdoor Graph}
\begin{figure}[t!]
\centering
    \includegraphics[scale=0.8]{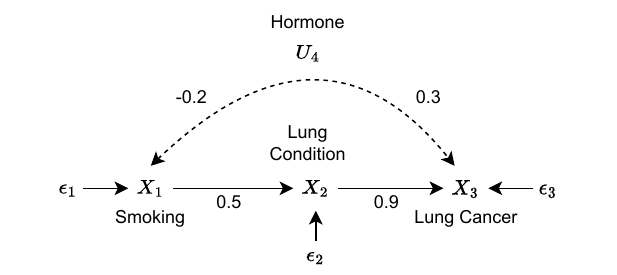}
    \caption{Frontdoor Causal Graph with edge weights}
    \label{fig:frontdoor-graph}
\end{figure}

We used following $\texttt{CGL-SCM}$ $M = \langle (\boldsymbol{U, \varepsilon}), \mathbf{X}, \mathbb{P}, \mathbf{F}_\mathbf{X} \rangle$ inducing a frontdoor graph as shown in Figure \ref{fig:frontdoor-graph} with $\mu = \begin{bmatrix} 0.3 & 0.1 & 0.2 \end{bmatrix}^T$.

\begin{table}[h]
\caption{Comparing Original and Estimated Causal Distributions for Frontdoor Graph}
\label{tab:frontdoor-table}
\vskip 0.15in
\begin{center}
\begin{small}
\begin{sc}
\begin{tabular}{lcccr}
\toprule
 & Original & Estimated \\
\midrule
$P(X_{3} | do(X_{2} = 1))$ & $\mathcal{N}(1.1, 1.09)$ & $\mathcal{N}(1.1018, 1.069)$ \\
$P(X_{3} | do(X_{1} = 1))$ & $\mathcal{N}(0.74, 1.9)$ & $\mathcal{N}(0.7391, 1.881)$ \\
\bottomrule
\end{tabular}
\end{sc}
\end{small}
\end{center}
\vskip -0.1in
\end{table}

\begin{table}[h]
\caption{Comparing Original and Estimated Causal Distributions for Napkin Graph}
\label{tab:napkin-table}
\vskip 0.15in
\begin{center}
\begin{small}
\begin{sc}
\begin{tabular}{@{}l@{\hspace{5pt}}c@{\hspace{5pt}}c@{}}
\toprule
 & Original & Estimated \\
\midrule
$P(X_{4} | do(X_{3} = 1))$ & $\mathcal{N}(0.3, 1.16)$ & $\mathcal{N}(0.3051, 1.1692)$ \\
$P(X_{4} | do(X_{1} = 1))$ & $\mathcal{N}(-1.068, 2.3248)$ & $\mathcal{N}(-0.9721, 2.3274)$ \\
\bottomrule
\end{tabular}
\end{sc}
\end{small}
\end{center}
\vskip -0.1in
\end{table}

\subsection{Napkin Graph}

We used following $\texttt{CGL-SCM}$ $M = \langle (\boldsymbol{U, \varepsilon}), \mathbf{X}, \mathbb{P}, \mathbf{F}_\mathbf{X} \rangle$ inducing a napkin graph as shown in Figure \ref{fig:napkin-graph} with $\mu = \begin{bmatrix} 0.8 & -0.9 & 0.01 & -0.5\end{bmatrix}^T$.





As we can see for both examples, estimated causal distributions are very close to original distributions. 
\section{Conclusion}

In this work, we introduced Centralized Gaussian Linear SCMs as a tractable subclass of Gaussian SCMs that preserve the identifiability of causal effects from observational data. We demonstrated their theoretical equivalence to GL-SCMs in terms of expressivity and identifiability and proposed an EM-based learning algorithm for parameter estimation, and then estimating causal effects once we know the parameters. Through experiments on canonical graphs like the frontdoor and napkin structures, we showed that our method accurately recovers interventional distributions from finite samples. Future works can explore more general distributions or finding bounds for non-identifiable queries from finite data.

\section{Impact Statements}

This paper presents work whose goal is to advance the field of Machine Learning. There are many potential societal consequences of our work, none which we feel must be specifically highlighted here.

\bibliography{example_paper}
\bibliographystyle{icml2025}

\newpage
\appendix
\onecolumn
\section{Background}
It is pivotal to introduce and elucidate the concepts of intervention and identifiability—crucial aspects in the realm of causality, integral to this paper.
\begin{definition}[Intervention]
Given two disjoint sets of variables in an SCM $M$, $\mathbf{Z}$, $\mathbf{Y}$ $\subseteq \mathbf{X}$, the causal effect of $\mathbf{Z}$ on $\mathbf{Y}$, denoted as $P(\mathbf{y} | do(\mathbf{z}))$, is a function from $\mathbf{Z}$ to the space of probability distributions of $\mathbf{Y}$. For each realization $\mathbf{z}$ of $\mathbf{Z}$, $P(\mathbf{y} | do(\mathbf{z}))$ gives the probability $\mathbf{Y} = \mathbf{y}$ induced by deleting from the model all equations corresponding to variables in $\mathbf{Z}$ and substituting $\mathbf{Z} = \mathbf{z}$ in the remaining equations.
\end{definition}

\begin{definition}[Identifiabilty]
Given the set of observed variables $\mathbf{X}$ such that $\mathbf{Z, Y} \subseteq \mathbf{X}$ the causal effect of $\mathbf{Z}$ on $\mathbf{Y}$ is said to be identifiable from a causal diagram $G$ if the quantity $P(\mathbf{y} \mid do(\mathbf{z}))$ can be computed uniquely from a positive probability of the observed variables. That is, if for every pair of models $M_1$ and $M_2$ inducing $G$, $$P^{M_1}(\mathbf{y} \mid do(\mathbf{z})) = P^{M_2}(\mathbf{y} \mid do(\mathbf{z}))$$ whenever $P^{M_1}(\mathbf{x}) = P^{M_2}(\mathbf{x}) > 0$.
\end{definition}
\section{Proofs}
\subsection{Proof of Theorem 2.3}
Let us take a $\texttt{GL-SCM}$ $M' = \langle (\boldsymbol{U', \varepsilon'}), \mathbf{X}, \mathbb{P}, \mathbf{F}_\mathbf{X} \rangle$. The endogenous variables $X_i$ are then given by
\[
X_i = \sum_{X_j \in Pa^o(X_i)} \alpha_{ji} X_j + \sum_{U'_k \in Pa^u(X_i)} \alpha_{ki}' U_k' + \mu_{i}' + \varepsilon_i'
\]

where $Pa^o(X_i)$ are the observed parents of $X_i$, and $Pa^u(X_i)$ are the unobserved confounding parents of $X_i$. $\mu_{i}'$ are the bias for $X_i$. $X_i$ can be re-written as:
\begin{align*}
X_i &= \sum_{X_j \in Pa^o(X_i)} \alpha_{ji} X_j + \sum_{U'_k \in Pa^u(X_i)} \alpha_{ki}' (\mu_{U'_k} + \sigma_{U'_k}^{2}U_k) +\mu_{i}' + \mu_{\varepsilon'_i} + \varepsilon_i \\
&= \sum_{X_j \in Pa^o(X_i)} \alpha_{ji} X_j + \sum_{U_k \in Pa^u(X_i)} \alpha_{ki} U_k + \mu_{i} + \varepsilon_i
\end{align*}
where,
\begin{gather*}
    \mathbf{U} \sim \mathcal{N}(\mathbf{0}, \mathbf{I}), \quad \boldsymbol{\varepsilon} \sim \mathcal{N}(\mathbf{0}, \boldsymbol{\Psi^2}), \quad
    \alpha_{ki} = \alpha_{ki}' \sigma_{U'_k}^{2}, \quad
    \mu_{i} = \Bigg(\sum_{U'_k \in Pa^u(X_i)} \alpha_{ki}' \mu_{U'_{k}}\Bigg) +\mu_{i}' + \mu_{\varepsilon'_i}
\end{gather*}
This is exactly the definition of a \texttt{CGL-SCM}. Thus for every \texttt{GL-SCM} $M'$ there exists a \texttt{CGL-SCM} $M$ such that $P^{M'}(\mathbf{X}) = P^{M}(\mathbf{X})$.

\subsection{Proof of Theorem 2.4}
The proof follows directly from the definition of identifiablity, which says that if a query $Q$ is identifiable from observational data $P(V)$ and causal graph, then all models which match on $P(V)$, must match on $Q$. Now, CGL-SCM is a subclass of GL-SCM, and if a $Q$ is identifiable from $P(V)$ and the causal graph, then it must also match in $Q$.
\section{Derivation of EM Steps}
The distributions for $\mathbf{U}$ and $\mathbf{X}$ can be written as \footnote{Assume that $\varepsilon$ has variance $\mathbf{I}$ for this part. Derivation for general $\boldsymbol{\Psi }$ is similar.}
\begin{gather*}
    \mathbf{U} \sim \mathcal{N}(\mathbf{0}, \mathbf{I}), \quad
    \boldsymbol{\varepsilon} \sim \mathcal{N}(\mathbf{0}, \mathbf{I}), \quad
    \mathbf{X} = B^T \boldsymbol{\mu} + B^TC^T \boldsymbol{U} + B^T \boldsymbol{\varepsilon}
\end{gather*}

From these, we can derive the join distribution of $(\mathbf{U}, \mathbf{X})$, after computing mean and variance

\[
\mathbb{E}[\mathbf{U}] = \mathbf{0}, \qquad \mathbb{E}[\mathbf{X}] = B^T\boldsymbol{\mu}
\]

Now, the variances $\Sigma_{UU} = I$ (from definition). $\Sigma_{UX}$ and $\Sigma_{XX}$ can be computed as follows:


\begin{align*}
\Sigma_{UX} &= \mathbb{E}[(\mathbf{U} - \mathbb{E}[\mathbf{U}])(\mathbf{X} - \mathbb{E}[\mathbf{X}])^T] =  \mathbb{E}[\mathbf{U}(B^T\bmu + B^TC^T\mathbf{U} + B^T\varepsilon - B^T\boldsymbol{\mu})^T] \\
&= \mathbb{E}[\mathbf{U}\mathbf{U}^TCB + \meps^TB] = \mE[\mU \mU^T] CB + \mE[\meps^T] B = CB
\end{align*}

Similarly, $\Sigma_{XX}$

\begin{align*}
\Sigma_{XX} &= \mE[(\mX - \mE[\mX])(\mX - \mE[\mX])^T] = \mE[(B^TC^T\mU + B^T\meps)(B^TC^T\mU + B^T\meps)^T] \\
&= \mE[B^TC^TUU^TCB + B^TC^T\mU\meps^TB + B^T\meps\mU^TCB + B^T\meps\meps^TB] \\
&= B^TC^T\mE[\mU \mU^T] CB + B^T \mE[\meps \meps^T] B = (CB)^TCB + B^TB
\end{align*}

The joint distribution $(\mU, \mX)$ can be written as
\[
\begin{bmatrix}
    \mU \\[2ex]
    \mX
\end{bmatrix}
\sim \mathcal{N}\left(
\begin{bmatrix}
    \mathbf{0} \\[2ex]
    B^T\bmu
\end{bmatrix},
\begin{bmatrix}
    \begin{aligned}
        &\mathbf{I} \\[2ex]
        &(CB)^T
    \end{aligned}
    \qquad
    \begin{aligned}
        &CB \\[2ex]
        &(CB)^TCB + B^TB
    \end{aligned}
\end{bmatrix}
\right)
\]

In the E-step, we compute the distribution of $\mU^i \mid \mathbf{x}^i; \bmu, B, C \sim \mathcal{N}\big(\bmu_{\mU^i \mid \mathbf{x}^i}, \Sigma_{\mU^i \mid \mathbf{x}^i}\big)$ where,
\begin{gather*}
    \bmu_{\mU^i \mid \mathbf{x}^i} = CB((CB)^TCB + B^TB)^{-1}(\mathbf{x}^i - B^T\bmu), \qquad
    \Sigma_{\mU^i \mid \mathbf{x}^i} = \mathbf{I} - CB((CB)^TCB + B^TB)^{-1}B^TC^T
\end{gather*}

The mean and variance of $\mX^i \mid \mathbf{u}^i; \bmu, B, C$ can be computed similarly,
\begin{gather*}
    \bmu_{\mathbf{X}^i \mid \mathbf{u}^i} = B^T\bmu + B^TC^T\mathbf{u}^i, \qquad \Sigma_{\mathbf{X}^i \mid \mathbf{u}^i} = B^TB
\end{gather*}

In the M-step, we want to maximize
\begin{align*}
&\sum_{i=1}^N \mE_{\mU^{i}|\mathbf{x}^{i}}[\log p(\mathbf{x}^i \mid \mU^i; \bmu, B, C)] \\
&= -\frac{nd}{2} \log(2\pi) - \frac{n}{2} \log|B^TB| - \frac{1}{2} \sum_{i=1}^N\mE\Big[\big(\mathbf{x}^i - B^T\bmu - B^TC^T\mU^i\big)^T\big(B^TB\big)^{-1}\big(\mathbf{x}^i - B^T\bmu - B^TC^T\mU^i\big)\Big] \\
\end{align*}

with respect to $B, C, \bmu$

In order to optimize this, we have to compute the third term and take expectation
\begin{align*}
&\sum_{i=1}^N\mE\Big[\big(\mathbf{x}^i - B^T\bmu - B^TC^T\mU^i\big)^T\big(B^TB\big)^{-1}\big(\mathbf{x}^i - B^T\bmu - B^TC^T\mU^i\big)\Big] \\
&= \sum_{i=1}^N(\mathbf{x}^i)^T\big(B^TB)^{-1}\mathbf{x}^i - 2(\mathbf{x}^i)^TB^{-1}\big(\bmu + C^T\mE_{\mU^i \mid \mathbf{x}^i}[\mU^i]\big) + \bmu^T\bmu + 2\bmu^TC^T\mE_{\mU^i \mid \mathbf{x}^i}[\mU^i] + \mE_{\mU^i \mid \mathbf{x}^i}\Big[\big(\mU^i\big)^T CC^T \mU^i\Big] \\
\end{align*}

The term $\mE_{\mU^i \mid \mathbf{x}^i}\big[(\mU^i)^TCC^T\mU^i\big]$ can be computed as follows:

\begin{align*}
\mE_{\mU^i \mid \mathbf{x}^i}\Big[\big(\mU^i\big)^T CC^T \mU^i\Big] &= \mE\Big[\sum_j \sum_{l,k} c_{lj} c_{jk} \cdot U_l^i U_k^i \mid \mathbf{x}^i\Big] = \sum_j \sum_{l,k} c_{lj}c_{jk} \mE[U_l^i U_k^i \mid \mathbf{x}^i] = \sum_{l,k} \sum_j c_{lj}c_{jk} \mE[U_l^i U_k^i \mid \mathbf{x}^i] \\
&= \sum_{l,k} \mE[U_l^i U_k^i \mid \mathbf{x}^i] \sum_j c_{lj}c_{jk} = \mE[\mathbf{U}^i{\mathbf{U}^i}^T \mid \mathbf{x}^i] \cdot CC^T = \big(\bmu_{\mU^i \mid \mathbf{x}^i}\bmu_{\mU^i \mid \mathbf{x}^i}^T + \Sigma_{\mU^i \mid \mathbf{x}^i}\big) \cdot CC^T
\end{align*}
\section{Experiments}\label{apx:experiments}
\subsection{Frontdoor Graph}
Following is the vectorized form of frontdoor $\texttt{CGL-SCM}$ $M = \langle (\boldsymbol{U, \varepsilon}), \mathbf{X}, \mathbb{P}, \mathbf{F}_\mathbf{X} \rangle$:

$\mathbf{U} = \begin{bmatrix}
    U_{4} \\
\end{bmatrix} \sim N(0, 1)$ \hspace{5 mm}
$\boldsymbol{\epsilon} = \begin{bmatrix}
    \epsilon_{1} \\
    \epsilon_{2} \\
    \epsilon_{3}
\end{bmatrix} \sim \mathcal{N}(0, \mathbf{I}')$ \hspace{5mm}
$\mu = \begin{bmatrix} 0.3 \\ 0.1 \\ 0.2 \end{bmatrix}$ \hspace{5mm}
$T = \begin{bmatrix}
    0 & 0.5 &  0 \\
    0 & 0 & 0.9 \\
    0 & 0 & 0
\end{bmatrix}$ \\ \\
$C = \begin{bmatrix}
    -0.2 & 0 &  0.3
\end{bmatrix}$ \hspace{5 mm}
$B = I + \sum_{i=1}^2 T^i = \begin{bmatrix}
    1 & 0.5 & 0.45 \\
    0 & 1 & 0.9 \\
    0 & 0 & 1
\end{bmatrix}$ \\ \\
$\mathbf{X} = \begin{bmatrix}
    X_{1} \\
    X_{2} \\
    X_{3}
\end{bmatrix} = B^T \boldsymbol{\mu} + B^TC^T \boldsymbol{U} + B^T \boldsymbol{\varepsilon}$ 

After running Alg.~\ref{alg-1} we obtained following estimations for $B, C$ and $\mu$ \\ \\
$\hat{\mu} = \begin{bmatrix} 0.3002 \\ 0.0963 \\ 0.2007\end{bmatrix}$ \hspace{5mm}
$\hat{C} = \begin{bmatrix}
    -0.2149 & 0 &  0.2626
\end{bmatrix}$ \hspace{5 mm}
$\hat{B} = \begin{bmatrix}
    1 & 0.5012 & 0.4491 \\
    0 & 1 & 0.9011 \\
    0 & 0 & 1
\end{bmatrix}$ \\ \\
After running algorithm 2 we obtained following estimation for $T$

$\hat{T} = \begin{bmatrix}
    0 & 0.5012 &  0 \\
    0 & 0 & 0.9011 \\
    0 & 0 & 0
\end{bmatrix}$
\subsection{Napkin Graph}
Following is the vectorized form of napkin $\texttt{CGL-SCM}$ $M = \langle (\boldsymbol{U, \varepsilon}), \mathbf{X}, \mathbb{P}, \mathbf{F}_\mathbf{X} \rangle$:

$\mathbf{U} = \begin{bmatrix}
    U_{5} \\
    U_{6}
\end{bmatrix} \sim N(0, \mathbf{I})$ \hspace{5mm}
$\boldsymbol{\epsilon} = \begin{bmatrix}
    \epsilon_{1} \\
    \epsilon_{2} \\
    \epsilon_{3} \\
    \epsilon_{4}
\end{bmatrix} \sim \mathcal{N}(0, \mathbf{I}')$ \hspace{5mm}
$\mu = \begin{bmatrix} 0.8 \\ -0.9 \\ 0.01 \\ -0.5\end{bmatrix}$ \hspace{5mm}
$T = \begin{bmatrix}
    0 & 0.1 &  0 & 0 \\
    0 & 0 & 0.9 & 0 \\
    0 & 0 & 0 & 0.8 \\
    0 & 0 & 0 & 0
\end{bmatrix}$ \\ \\ \\
$C = \begin{bmatrix}
    -0.2 & 0 &  0.1 & 0 \\
    0.3 & 0 & 0 & 0.4
\end{bmatrix}$ \hspace{5 mm}
$B = I + \sum_{i=1}^3 T^i = \begin{bmatrix}
    1 & 0.1 & 0.09 & 0.072 \\
    0 & 1 & 0.9 & 0.72 \\
    0 & 0 & 1 & 0.8 \\
    0 & 0 & 0 & 1
\end{bmatrix}$ \\ \\
$\mathbf{X} = \begin{bmatrix}
    X_{1} \\
    X_{2} \\
    X_{3} \\
    X_{4}
\end{bmatrix} = B^T \boldsymbol{\mu} + B^TC^T \boldsymbol{U} + B^T \boldsymbol{\varepsilon}$ 

After running Algorithm \ref{alg-1} we obtained following estimations for $B, C$ and $\mu$ \\ \\ 
$\hat{\mu} = \begin{bmatrix} 0.7994 \\ -0.9044 \\ 0.0113 \\ -0.4894\end{bmatrix}$ \hspace{5mm}
$\hat{C} = \begin{bmatrix}
    -0.1271 & 0 &  0.1665 & 0 \\
    0.3389 & 0 & 0 & 0.4113
\end{bmatrix}$\hspace{5mm}
$\hat{B} = \begin{bmatrix}
    1 & 0.1024 & 0.0920 & 0.0566 \\
    0 & 1 & 0.8985 & 0.7134 \\
    0 & 0 & 1 & 0.7945 \\
    0 & 0 & 0 & 1
\end{bmatrix}$ \\ \\
After running algorithm 2 we obtained following estimation for $T$

$\hat{T} = \begin{bmatrix}
    0 & 0.1024 &  0 & 0 \\
    0 & 0 & 0.8985 & 0 \\
    0 & 0 & 0 & 0.7945 \\
    0 & 0 & 0 & 0
\end{bmatrix}$


\end{document}